# Time-Optimal Path Tracking for Industrial Robots: A Dynamic Model-Free Reinforcement Learning Approach

Jiadong Xiao, Lin Li, Tie Zhang, and Yanbiao Zou

*Abstract*—In pursuit of the time-optimal path tracking (TOPT) trajectory of a robot manipulator along a preset path, a beforehand identified robot dynamic model is usually used to obtain the required optimal trajectory for perfect tracking. However, due to the inevitable model-plant mismatch, there may be a big error between the actually measured torques and the calculated torques by the dynamic model, which causes the obtained trajectory to be suboptimal or even be infeasible by exceeding given limits. This paper presents a TOPT-oriented SARSA algorithm (TOPTO-SARSA) and a two-step method for finding the time-optimal motion and ensuring the feasibility : Firstly, using TOPTO-SARSA to find a safe trajectory that satisfies the kinematic constraints through the interaction between reinforcement learning agent and kinematic model. Secondly, using TOPTO-SARSA to find the optimal trajectory through the interaction between the agent and the real world, and assure the actually measured torques satisfy the given limits at the last interaction. The effectiveness of the proposed algorithm has been verified through experiments on a 6-DOF robot manipulator.

## I. Introduction

THE research of the time-optimal path tracking (TOPT) for robotic manipulators can be dated back as far as the early 1970 s [1] . The goal is to find a feasible optimal trajectory which satisfies the constraint conditions along a given path. The optimal objective can be minimal consumed energy or execution time. The solution to this problem is a mapping of the geometric path to a time-dependent trajectory, where the required feasibility implies that the given constraint bounds such as torque or joint velocity/acceleration bound must be met. Since the mention of the TOPT problem, a variety of approaches to find the optimal solution for the robot manipulators have been proposed, such as dynamic trajectory scaling [2] , integrating the maximum acceleration in order to obtain bang-bang acceleration profiles [3-5], using the dynamic programming to find a trajectory that minimizes cost [6-8] , formulating the TOPT problem as a convex optimization problem and subsequent solving it by using convex optimization toolbox [9-11].

In all of the researches as mentioned above, the optimal trajectories are all obtained from maximizing the calculated torques by the dynamic model $\tau_{model}$ within the torque constraints. However, as there is an inevitable model-plant mismatch, the dynamic model does not reflect the real situation, and the calculated torques by the dynamic model $\tau_{model}$ have a large difference from the actually measured torque $\tau_{measured}$. In most cases, although $\tau_{model}$ are within the torque constraint limit, $\tau_{measured}$ may also exceed the torque constraint limit, which causes the servo motor to operate under overload conditions, reduces the life of the motor, and even causes unexpected downtime. To tackle the problem of model-plant mismatch, most of the researches adopt the solutions: increasing the complexity of the dynamic model to make $\tau_{model}$ closer to $\tau_{measured}$: reaching from considering friction effects [10] to adding the iterative compensation term to compensate the model error [11, 12]. But even so, there is still no dynamic model which can reflect the real situation completely. Therefore, it is necessary to jump out of the orbit of traditional thinking to find a new method that can avoid the model-plant mismatch phenomenon and obtain the optimal trajectory.

Inspired by the idea of using reinforcement learning (RL) for vehicle time-optimal velocity control in [13], we think that reinforcement learning can be an effective method in solving the TOPT problem of industrial robots. Reinforcement learning is a computational approach to understanding and automating goal-directed learning and decision-making [14]. It is inspired by the trial-and-error learning process related to the dopaminergic system [15]. RL is distinguished from other computational approaches by its emphasis on learning by an RL agent from direct interaction with its environment, without relying on exemplary supervision or complete models of the environment. Through the experience obtained from the interaction between the RL agent and environment, the RL model aims to maximize rewards and minimize penalties.

Since the concept of reinforcement learning was first proposed in the engineering literature in the 1960 S [16] , a variety of reinforcement learning approaches have been

This work is supported by National Science and Technology Major Project of China (No. 2015ZX04005006), Science and Technology Planning Project of Guangdong Province, China (2015B010918002), Science and Technology Major Project of Zhongshan city, China (No. 2016F2FC0006, 2018A10018.

J. Xiao, L. Li, T. Zhang and Y. Zou are with the school of mechanical and automotive engineering, South China University of Techonology, Guangzhou 510641 (e-mail: mejacktonshaw@mail.scut.edu.cn; linli@scut.edu.cn; merobot@scut.edu.cn; ybzou@scut.edu.cn)

proposed. Reinforcement learning was originally used in the disciplines of game theory, information theory, control theory, and operation research. With the development of reinforcement learning theory, it has been adopted in the field of robotic control, and has a large number and variety of applications, such as path planning of mobile robot [17], gait generation for robots [18], obstacle avoidance of robot manipulator [19], and robotic assembly [20]. Although model-based reinforcement learning algorithm does exist, most of the above-mentioned reinforcement learning algorithms for robots are model-free.

SARSA, a typical model-free reinforcement learning algorithm, due to its simplicity and requiring less computational power [14], has been widely used in the field of robotics [21-24]. Therefore, we are able to find the time-optimal trajectory without considering the robot dynamic model by using SARSA. SARSA uses the idea of exploration and exploitation. At the beginning of the RL process, we use the exploration to acquire RL experience. At the end of the RL process, we use the exploitation to obtain a policy which maximizes the long term return.

Although both subjects—SARSA and the TOPT problem of industrial robots—attracted wide attention in the past and are still actively researched, their combination remains rare due to their inherently unlike nature: SARSA is mostly used to find the minimum steps for the agent to reach the target state, such as Windy Gridworld and pole-balancing task[14]. However, the TOPT problem aims to find a trajectory which has the greatest velocity of each point on the path.

The goal of this research is to reformulate the TOPT problem as a reinforcement learning problem which can be solved by using SARSA and to obtain a time-optimal trajectory which has the maximum $\tau_{measured}$ within the given limit. The rest of this paper is structured as follows: The TOPT problem in the grid-based map is defined in Section II. Section III briefly reviews the SARSA algorithm and further explains how it is applied to the TOPT problem. Section IV presents a TOPT-oriented SARSA algorithm (TOPTO-SARSA) and a two-step method to solve the TOPT problem. Section V presents the experimental evaluation of our methods on a 6-DOF robot manipulator. Finally, Section VI concludes this research and addresses future works.

## II. PROBLEM STATEMENTS AND ENVIRONMENT DESCRIPTION

### A. Problem Statements

Since the TOPT problem was introduced in the early 1970 s, most of the methods to tackle the TOPT problem aimed to optimize the scalar function $t \to s(t)$ [4, 9, 25]. In this scalar function, $s \in [0,1]$ is the pseudo-displacement, which represents the "position" on the path at each time instant.

The optimization goal of the TOPT problem is to minimize the trajectory execution time. Thus, the optimization objective function can be expressed as

$$\min T = \int_0^T 1 dt \quad (1)$$

We rewritten the objective function (1) by changing the integration variable from $t$ to $s$, as follow

$$\min T = \int_0^T 1 dt = \int_{s(0)}^{s(T)} 1/\dot{s}\, ds = \int_0^1 1/\dot{s}\, ds \quad (2)$$

where $\dot{s} = ds/dt$ is the pseudo-velocity.

Therefore, the TOPT problem can be transformed into the planning problem for the pseudo-velocity $\dot{s}$ in the phase plane $s - \dot{s}$. The optimization objective is to seek an optimal trajectory which starts from the initial state (0,0), ends at the terminate state (1,0), and has the maximum pseudo-velocity $\dot{s}$ limited by the dynamic or kinematic constraints in the phase plane $s - \dot{s}$ [4, 5, 25].

### B. Environment Description

The above-mentioned TOPT problem is similar to the mobile robot optimal path planning problem which aims to search an optimal path to avoid all the obstacles and reach the objective place as soon as possible [17, 26], as shown in Fig. 1. However, there is also a difference between the two cases: the TOPT problem aims to maximize the pseudo-velocity $\dot{s}$ instead of minimizing the steps for reaching the objective place.

The planning environment for the TOPT problem is a grid-based two-dimensional (2-D) field with a continuous curved obstacle in it, as shown in Fig. 2. The two dimensions are pseudo-displacement $s$ and pseudo-velocity $\dot{s}$. The obstacle is the infeasible area related to the maximum velocity curve (MVC, see[4]) which obtained based on the dynamic and kinematic constraint conditions. Hence, the grid is divided into the feasible area and the infeasible area. The start place is (0,0) and the objective place is (1,0).

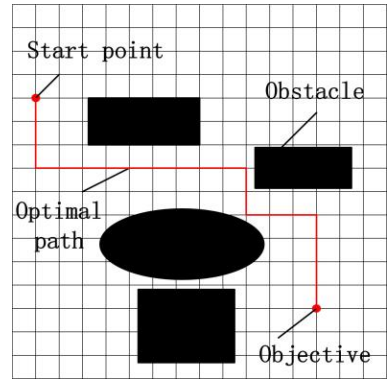

Fig. 1. Illustration of mobile robot optimal path planning

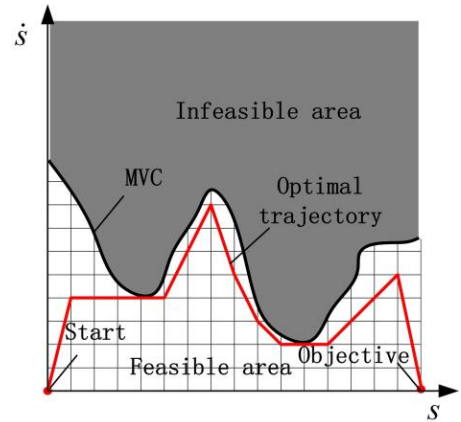

Fig. 2. Illustration of TOPT

## III. SARSA AND ITS APPLICATION ON TOPT

*A. Briefly Review of SARSA*

State–action–reward–state–action (SARSA) is an algorithm for learning a Markov decision process policy, used in the reinforcement learning area of machine learning. It was proposed by Rummery and Niranjan in a technical note with the name "Modified Connectionist Q-Learning" (MCQ-L)[27]. The alternative name SARSA (as is known to us), was proposed by Rich Sutton [14].

The general form of the SARSA algorithm is shown in Algorithm 1, where the action-value function is as follows:
$$Q(S_k, A_k) \leftarrow Q(S_k, A_k) + \alpha[R_{k+1} + \gamma Q(S_{k+1}, A_{k+1}) - Q(S_k, A_k)] \quad (3)$$

where：

$S_k$ is the current state;
$A_k$ is the action performed in the $S_k$;
$S_{k+1}$ is the next state;
$R_{k+1}$ is the reward or penalty received from the environment when the agent takes the action $A_k$ in state $S_k$;
$A_{k+1}$ is the action performed in the $S_{k+1}$ state;
$\gamma$ is the discount factor ($0 \leq \gamma < 1$);
$\alpha$ is the learning coefficient ($0 < \alpha < 1$).

The $\varepsilon - greedy$ algorithm as follows[14]:
$$A_{k+1} \leftarrow \begin{cases} \arg max_{A_{k+1}} Q(S_{k+1}, A_{k+1}), with\ probability\ 1 - \varepsilon\ (exploit) \\ a\ random\ action,\quad with\ probability\ \varepsilon\ (explore) \end{cases} \quad (4)$$

---

**Algorithm 1:** Classical SARSA algorithm

1: Initialize $Q(S, A), \forall S \in \mathbb{S}, A \in \mathbb{A}(S)$, arbitrarily, and $Q(terminal - state,\cdot) = 0$
2: **while** ( iteration<Max iteration)
3:   Select a starting state, $Q(S_1, A_1)$
4:   $k = 1$
5:   **while** goal is not achieved
6:     Choose $A_k$ from $S_k$ using policy derived from Q (e.g., $\varepsilon - greedy$)
7:     Take action $A_k$, observe $R_{k+1}, S_{k+1}$
8:     Choose $A_{k+1}$ from $S_{k+1}$ using policy derived from Q (e.g., $\varepsilon - greedy$)
9:     $Q(S_k, A_k) \leftarrow Q(S_k, A_k) + \alpha[R_{k+1} + \gamma Q(S_{k+1}, A_{k+1}) - Q(S_k, A_k)]$
10:    $S_k \leftarrow S_{k+1}; A_k \leftarrow A_{k+1}; k \leftarrow k + 1$
11:  **end while**
12: **end while**
13: **end**

---

*B. Application*

To use SARSA to solve the TOPT problem, the phase plane $s - \dot{s}$ needs to be divided into a discrete grid. The discretization of the $s$-dimension is the same as path discretization. The path discretization preprocess has been used in other methods [8, 9, 25, 28]. As in the convex-optimization-based approach of [9, 28], we divide the pseudo-displacement interval [0,1] into $N - 1$ segments and $N$ grid points:

$$0 =: s_0, s_1, \ldots, s_{N-2}, s_{N-1} := 1$$

As in the approach of [2], we divide the pseudo-velocity interval $[0, \dot{s}_{MVC}]$ into $M - 1$ segments and $M$ grid points:

$$0 =: \dot{s}_0, \dot{s}_1, \ldots, \dot{s}_{M-2}, \dot{s}_{M-1} := \dot{s}_{MVC}$$

Hence, the phase plane $s - \dot{s}$ is divided into an $N \times M$ grid. In this grid, a grid point $(s_k, \dot{s}_k)$ is a state $S_k$, and a movement between the current state $S_k$ and the next state $S_{k+1}$ is an action; therefore, an action of state $S_k$ equals the next state $S_{k+1}$.

To avoid the measured torques or accelerations exceeding the dynamic or kinematic constraints, the grid points not satisfying the constraint conditions are set to infeasible states. Moreover, to make the optimal trajectory end at (1,0), all grid points on the line $s = 1$ except for (1,0) are set to infeasible states, as shown in Fig. 3.

To obtain an optimal trajectory with the maximum pseudo-velocity $\dot{s}$, the RL agent must be given rewards and penalties to regulate the learning behavior. As shown in Fig. 3, in the feasible area, heading to the upper area means greater rewards; simultaneously, the probabilities of suffering penalties increase as the agent moves closer to the infeasible states. Therefore, the reward and penalty of RL should be related to the pseudo-velocity, established as follows:

$$R_{k+1} = \begin{cases} \dot{s}_k + \dot{s}_{k+1} & reward \\ -(\dot{s}_k + \dot{s}_{k+1}) & penalty \end{cases} \quad (5)$$

In addition, in contrast to the mobile robot optimal path planning problem (which only gives the agent four actions (up, down, left, or right) to be taken), the TOPT problem gives the agent more actions to be taken if these actions are within the action range. The action range is calculated according to current state of the agent and constraint conditions (such as dynamic or kinematic constraints). Moreover, the closer the RL agent is to the MVC, the fewer feasible actions that can be selected [5]. When the agent is on the MVC, there is only one feasible action since the maximum pseudo-acceleration is equal to the minimum pseudo-acceleration (sometimes, there is no feasible action if the calculated pseudo-velocity of the next state is not on a grid point). Finally, to improve the computational efficiency, all actions with a Q value less than 0 within the action range should not be taken, as an action with such a Q value will cause the agent to reach an infeasible state. The analysis is as follows: As shown in the upper diagram of Fig. 4, the closer the agent is to the constraint boundary, the smaller the range of actions. It is true that some states close to the MVC are within the feasible area and satisfy the constraint conditions; however, when the agent is moving forward from these states with the minimum acceleration, it will inevitably reach the infeasible area, resulting in an unsuccessful episode. We call these states critical states. The initial Q value of these critical states is greater than 0; however, as the learning process continues, by using the action-value function to update the Q value, the Q value will gradually decrease and eventually become less than 0. As shown in the diagram on the lower part of Fig. 4, if the agent keeps away from these critical states, the agent will acquire an optimal trajectory that does not violate the constraints. In addition, when the agent is in a special state, the Q value of all feasible actions of that state is less than 0, which means that the agent is directed to the critical state and will inevitably reach an infeasible area. Therefore, these special states should also be regarded as critical states and penalize the agent.

Hence, we can use SARSA to solve the TOPT problem, with the algorithm as follows.

| **Algorithm 2:** SARSA for solving the TOPT problem |
|---|
| 1:    Discrete the phase plane $s - \dot{s}$ into a $N \times M$ grid |
| 2:    Initialize $Q(S, A), \forall S \in \mathbb{S}, A \in \mathbb{A}(S)$, arbitrarily, and $Q(terminal - state, \cdot) = 0$ |
| 3:    **while** ( iteration<Max iteration) |
| 4:      Select a starting state (0,0) |
| 5:      $k = 1$ |
| 6:      Calculate the range of action $\mathbb{A}(S_k)$ |
| 7:      Choose $A_k \in \mathbb{A}(S_k)$ from $S_k$ using policy derived from $Q$ (e.g., $\varepsilon$ − greedy) |
| 8:      **while** $k \leq N$ & $R_{k+1} \geq 0$ |
| 9:        Take action $A_k$, observe $R_{k+1}, S_{k+1}$ |
| 10:       Calculate the range of action $\mathbb{A}(S_{k+1})$ |
| 11:       **If** is empty($\mathbb{A}(S_{k+1})$) |
| 12:         $Q(S_{k+1}, A_{k+1}) = 0$ |
| 13:       Choose $A_{k+1} \in \mathbb{A}(S_{k+1})$ from $S_{k+1}$ using policy derived from $Q$ (e.g., $\varepsilon$ − greedy) |
| 14:       $Q(S_k, A_k) \leftarrow Q(S_k, A_k) + \alpha[R_{k+1} + \gamma Q(S_{k+1}, A_{k+1}) - Q(S_k, A_k)]$ |
| 15:       $S_k \leftarrow S_{k+1}; A_k \leftarrow A_{k+1}; k \leftarrow k + 1$ |
| 16:      **end while** |
| 17:    **end while** |
| 18:    **end** |

## IV. TOPT-Oriented SARSA Algorithm (TOPTO-SARSA)

### A. TOPTO-SARSA

Although the SARSA algorithm can be used to solve the robotic time-optimal path tracking problem, it has some limitations that may affect the problem-solving efficiency.

First, when selecting an action and using the $\varepsilon$-greedy algorithm, the action with the greatest Q value has the greatest probability to be selected. If the action with the greatest Q value is selected and taken, and after interacting with the environment, the agent will not touch the infeasible area; then, the selected action's Q value will be greater. This creates a loop. On the one hand, for an action, a greater Q value will make it more likely to be selected than other actions, which may result in an increase in the number of times it is selected. On the other hand, this increased number of times the action is selected will increase the action's Q value. Such a loop will make the program become stuck in local optima, leading to an increase in computation time and degrading the quality of the solution. To avoid such limitations, in the exploitation mode of the $\varepsilon$-greedy algorithm, the selected action should be the action with the greatest pseudo-velocity instead of the greatest Q value in the action range. In addition, when the agent reaches the previous single state of the termination state (1,0), if the termination state is within the feasible action range of the last action, the termination state (1,0) should be selected (as there is only one termination state; the other states are redundant.). The improved $\varepsilon$-greedy algorithm is expressed as follows:

$$A_{k+1} \leftarrow \begin{cases} \arg max_{A_{k+1}} A_{k+1}, \text{ with probability } 1 - \varepsilon \text{ (exploit)} \\ a \text{ random action, with probability } \varepsilon \text{ (explore)} \\ (1,0), \text{ if } (1,0) \in terminate \text{ action range} \end{cases} \quad (6)$$

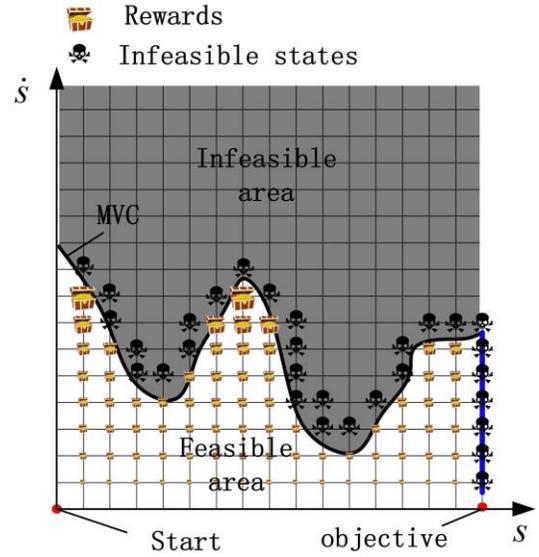

Fig. 3. Illustration of rewards and infeasible states

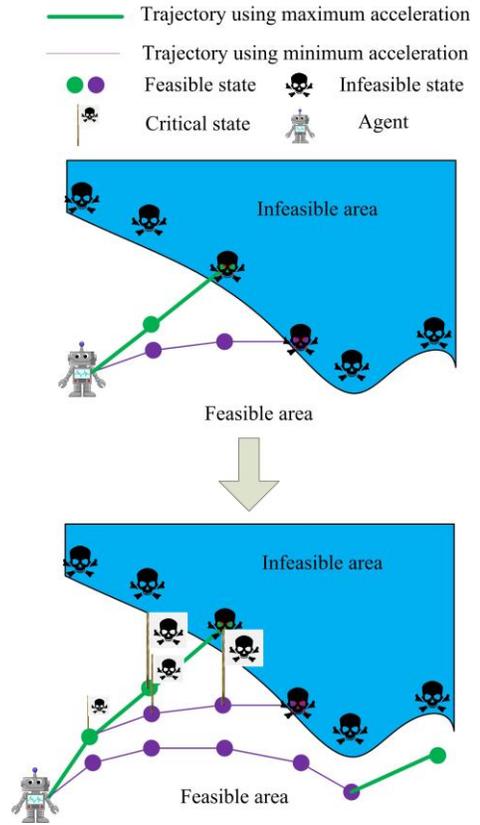

Fig. 4. Illustration of critical states

Second, since SARSA is a one-step update algorithm, when an agent reaches an infeasible area and receives a penalty, it can only propagate this penalty back to the previous states through a single-step update. Therefore, it takes a significant amount of time to update to make the Q value of the critical states less than 0 for ensuring that the agent avoids reaching these states. By speeding up this process by adding a penalty term in the action-value function, all states of an unsuccessful episode can learn the experience of a failure, thus accelerating the process. The improved action-value function is expressed as follows:

$$Q(S_k, A_k) \leftarrow Q(S_k, A_k) + \alpha[R_{k+1} + \gamma Q(S_{k+1}, A_{k+1}) - Q(S_k, A_k)] + \rho^{K-k} R_{K+1} \quad (7)$$

where K is the total number of states in an unsuccessful episode, $R_{K+1}$ is the penalty that the agent receives when it reaches an infeasible state, and $\rho (0 < \rho < 1)$ is the penalty discount factor.

Finally, to improve the computational efficiency, after a successful episode, we set the greed factor ε to 0 to exploit the learning experience to obtain the optimal policy, save the optimal policy, and then re-set the greed factor $0 < \varepsilon < 1$ to explore again. After another successful episode, we reset the greed factor ε to 0 to exploit the learning experience and obtain another optimal policy. If the newly obtained policy is equal to the previous optimal policy, the agent may have traversed all possible situations, and the algorithm converges to the optimal policy.

The TOPTO-SARSA algorithm is given in Algorithm 3.

| **Algorithm 3:** TOPTO-SARSA |
|---|
| 1:    Discrete the phase plane $s - \dot{s}$ into a $N \times M$ grid |
| 2:    Initialize $Q(S, A), \forall S \in \mathbb{S}, A \in \mathbb{A}(S)$, arbitrarily, and $Q(terminal-state, \cdot) = 0$ |
| 3:    **while** ( iteration<Max iteration) |
| 4:      Select a starting state (0,0) |
| 5:      $k = 1$ |
| 6:      Calculate the range of action $\mathbb{A}(S_k)$ |
| 7:      Choose $A_k \in \mathbb{A}(S_k)$ from $S_k$ using policy derived from $Q$ (e.g., improved ε − greedy) |
| 8:      **while** $k \leq N$ & $R_{k+1} \geq 0$ |
| 9:        Take action $A_k$, observe $R_{k+1}, S_{k+1}$ |
| 10:       Calculate the range of action $\mathbb{A}(S_{k+1})$ |
| 11:       **If** $R_{k+1}$<0 or is empty ($\mathbb{A}(S_{k+1})$) |
| 12:         **If** is empty ($\mathbb{A}(S_{k+1})$) |
| 13:           $Q(S_{k+1}, A_{k+1}) = 0$ |
| 14:           $Q(S_k, A_k) \leftarrow Q(S_k, A_k) + \alpha[R_{k+1} + \gamma Q(S_{k+1}, A_{k+1}) - Q(S_k, A_k)]$ |
| 15:         **For** $i = 1, \dots, k$ **do** |
| 16:           $Q(S_k, A_k) \leftarrow Q(S_k, A_k) + \rho^{k-i} R_{k+1}$ |
| 17:         break |
| 18:       **Else** |
| 19:         Choose $A_{k+1} \in \mathbb{A}(S_{k+1})$ from $S_{k+1}$ using policy derived from $Q$(e.g., improved ε-greedy) |
| 20:         $Q(S_k, A_k) \leftarrow Q(S_k, A_k) + \alpha[R_{k+1} + \gamma Q(S_{k+1}, A_{k+1}) - Q(S_k, A_k)]$ |
| 21:         $S_k \leftarrow S_{k+1}; A_k \leftarrow A_{k+1}; k \leftarrow k+1$ |
| 22:      end while |
| 23:      **If** $k = N$ |
| 24:        **If** $\varepsilon > 0$ |
| 25:          let $\varepsilon = 0$ to exploitation and obtained the optimal policy |
| 26:        **else** |
| 27:         **If** the optimal policy is updated |
| 28:           Save the optimal policy and initialize ε to explore |
| 29:         **else** |
| 30:           break |
| 31:    end while |
| 32:    end |

### B. Two Step Method

Since the dynamic model is not taken into account in this paper, the action range can not be calculated. In this case, the initial trajectory is generated randomly, and the servo motor may be damaged if we run this trajectory. In order to avoid this, the acceleration should be limited to avoid placing an excessive load on the motor.

Assume that the displacement of each joint is a path $\mathbf{q}(s)$ which is a function of the pseudo-displacement $s$, whereas the pseudo-displacement $s = s(t)$ is a scalar function of time $t$. Without loss of generality, it is assumed that the trajectory starts at $t = 0$, ends at $t = T$ and that $s(0) = 0 \leq s(t) \leq 1 = s(T)$. Differentiating $\mathbf{q}(s(t))$ with respect to t yields

$$\dot{\mathbf{q}}(s) = \mathbf{q}'(s)\dot{s} \quad (8)$$
$$\ddot{\mathbf{q}}(s) = \mathbf{q}'(s)\ddot{s} + \mathbf{q}''(s)\dot{s}^2 \quad (9)$$

Considering the given acceleration limits $\ddot{\mathbf{q}}_{min}$ and $\ddot{\mathbf{q}}_{max}$, combined with equation (9), yields

$$(\ddot{\mathbf{q}}_{min} - \mathbf{q}''(s)\dot{s}^2)/\mathbf{q}'(s) \leq \ddot{s} \leq (\ddot{\mathbf{q}}_{max} - \mathbf{q}''(s)\dot{s}^2)/\mathbf{q}'(s) \quad (10)$$

Each row $i$ of (10) is of the form

$$(\ddot{q}_{i,min} - q_i''(s)\dot{s}^2)/q'_i(s) \leq \ddot{s} \leq (\ddot{q}_{i,max} - q_i''(s)\dot{s}^2)/q_i'(s) \quad (11)$$

Thus, the maximum/minimum pseudo-acceleration can be obtained as

$$\ddot{s}_{max,k} = \min_i((\ddot{q}_{i,max} - q_i''(s)\dot{s}^2)/q_i'(s)) \quad (12)$$
$$\ddot{s}_{min,k} = \max_i((\ddot{q}_{i,min} - q_i''(s)\dot{s}^2)/q'_i(s)) \quad (13)$$

The feasible action range of the agent on the state $S_k = (s_k, \dot{s}_k)$ is $A_k \in (\ddot{s}_{min,k}, \ddot{s}_{max,k})$. When the minimum pseudo-acceleration $\ddot{s}_{min,k}$ of a certain state is greater than the maximum pseudo-acceleration $\ddot{s}_{max,k}$, the state is infeasible, and when the agent reach this state, it should be got a penalty.

Hence, we can obtain the initial trajectory for interacting and the specialized initial Q value by setting the feasible action range and the infeasible state and using the improved SARSA algorithm for learning. Moreover, since Q value has been specialized after obtaining the initial trajectory, it will not take a lot of time to explore in subsequent learning.

Using the TOPTO-SARSA to interact with the real world environment mainly includes the following two steps:

Step 1. Run the obtained initial trajectory (or the optimal trajectory obtained from subsequent learning) on the robot manipulator and obtain $\tau_{measured}$.

Step 2. If the corresponding $\tau_{measured}$ of any discrete point in the trajectory exceeds the constraint limit, the corresponding state is set to an infeasible state. Hence, the agent will not pass through this state in the next exploration. After setting the infeasible states, restarting the learning process to obtain a new optimal trajectory and updating the specialized Q-value. And then return to step 1 to run the optimal trajectory.

Repeat steps 1 and 2 until there is no $\tau_{measured}$ exceed the torque constraint limit.

## V. EXPERIMENTAL VALIDATION

### A. Experiment Condition

*Experiment setting*

The experimental setup is shown in Fig. 5. It consists of 4 parts: *1. Industrial robot.* All experiments are implemented in the GSK-RB03A1 6-DOF industrial robot of Guangzhou CNC Equipment Co., Ltd. *2. Servo drivers.* The robot is driven by CoolDrive R series alternating current (AC) servo drivers, which receive the control commands from the industrial control personal computer (PC) and send them to the servo motor in addition to receiving the encoder values and current values returned by the servo motor in real time and sending them to the industrial control PC to calculate $\tau_{measured}$. *3. Industrial control PC.* The industrial control PC used in this paper is a DT-610P-ZQ170MA industrial PC with an Intel Core i7-4770 3.4 GHz eight-core processor, 8 GB memory, and Windows 7 64-bit system, which is used to plan the trajectory in MATLAB R2018b and execute the trajectory in the control software platform. *4. Control software platform.* The control software platform is mainly constructed based on a Windows 7 64-bit system and a real-time kernel control system. The EtherCAT Industrial Ethernet bus protocol is adopted for communication between the control platform and the servo drivers, with a 1-ms control cycle.

*Reinforcement learning parameters setting*

The discount factor $\gamma$, learning coefficient $\alpha$ and penalty discount factor $\rho$ are all set to 0.8. The greed factor $\varepsilon$ of the greedy algorithm used for exploration is set to 0.4. The maximum number of episodes is set to 100000.

*Experiment path*

A line path and a curved path are used to verify the feasibility of the algorithm, where the line path is a straight line with starting point (166.8,-464.7,132.3)(mm) and ending point (259.5,420.1,132.3)(mm), and the curved path is a cosine curved with expression $x = 350 + 150\cos(16y)$, $y \in [-300,300]$, $z = 40$(mm), as shown in Fig.6.

*Grid division*

We use the method of [9] and [8] to divide the phase plane $s - \dot{s}$ into a 350×500 grid for the line path case and 550×500 grid for the cosine curved path case.

*Comparative experiment*

To verify the effectiveness of the algorithm, the dynamic model-based direct planning method NI-like is chosen as the comparative algorithm as it also considers the case of acceleration constraints.

*Constraint conditions*

The joint acceleration constraints are set to $A_{max/min} = \pm[50; 42; 70; 80; 80; 80](rad/s^2)$. The torque constraints are set to. $\tau_{max/min} = \pm[104; 103; 35; 8; 6.5; 6.5](N \cdot m)$.

### B. Performance Comparison between SARSA and TOPTO-

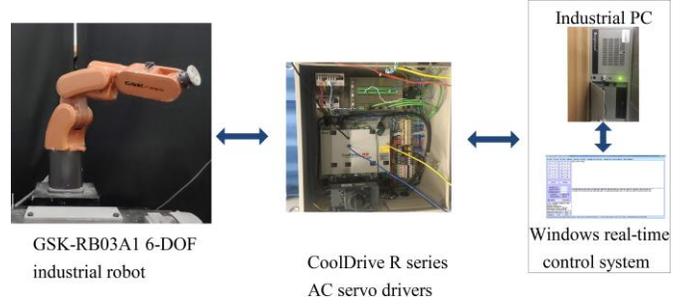

Fig. 5. Experiment settings

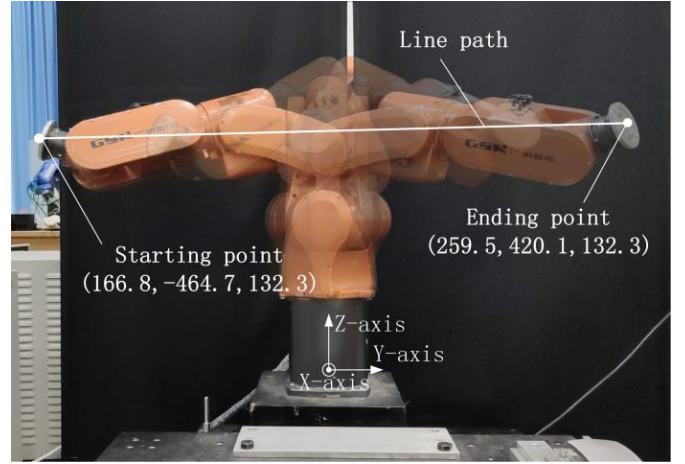

(a)

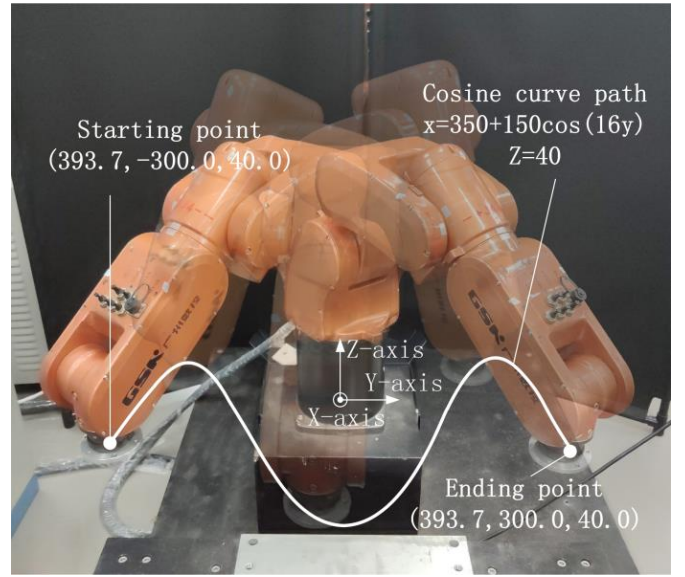

(b)
Fig. 6. Experiment paths: (a) Line path; (b) Cosine curved path

*SARSA*

To illustrate the superior effect of the TOPTO-SARSA algorithm relative to the SARSA algorithm, we compare the performance between the two algorithms. The two algorithms are used to obtain the initial interaction trajectory as mentioned in Section IV-B.

*Convergence*: We use the trajectory execution time (which is our optimization objective and it is obtained by exploiting the RL experience of the successful episodes.) to judge whether the successful episodes converge to the optimal episode. To avoid the influence of irrelevant variables, TOPTO-SARSA algorithm does not stop early but continues until the episode number reaches the maximum limit. Fig. 7 shows the trajectory execution time by using the two algorithms (to make it clear, we only shows the trajectory time of the successful episodes, while the time of unsuccessful episodes is not shown). From the result of Fig.7, we find that by using TOPTO-SARSA, once we get a successful episode, the episode is the optimal episode. Whereas by using SARSA, the algorithm still not converge to the optimal episode even the episode number reaches the maximum episode limit, as the original epsilon greedy algorithm is not suitable for solving the TOPT problem.

*Trajectory execution time*: As shown in Fig. 7, the optimal trajectory execution time by using TOPTO-SARSA is apparently less than that by using SARSA.

*Computational time: Line path case*: By using TOPTO-SARSA, we need 14.537 s to wait for the first successful episode, whereas by using SARSA, we need 30.104 s to wait for the first successful episode. *Cosine curve path case*: By using TOPTO-SARSA, we need 18.322 s to wait for the first successful episode, whereas by using SARSA, we need 35.937 s to wait for the first successful episode. Moreover, by using SARSA, it takes significant time for us to wait for the episode number reaches the maximum limit whereas the algorithm still can not converge to the optimal policy.

### C. Experimental Evaluation of the Two-Step TOPTO-SARSA Method

As the higher efficiency of TOPTO-SARSA has been verified in Section V-B, in this section, we validate the effectiveness of TOPTO-SARSA through experiments on a 6-DOF industrial robot.

*Case 1 Line*

*Computational time*: the computational time for NI-like is 1.045 s, whereas the computational time for learning the initial trajectory is 14.537 s. However, after the learning of the initial trajectory, since the Q value has been specialized, the computational time for each learning is less than 0.1 s.

*Trajectory execution time:* the trajectory execution time by using NI-like is 0.8004 s, whereas the trajectory execution time of the initial trajectory by using TOPTO-SARSA is 0.7192s and after 55 interactions, the trajectory execution time is 0.7806s, which has a 2.4% reduction in trajectory execution time compared to use NI-like. Moreover, by using TOPTO-SARSA, $\tau_{measured}$ is no longer exceed the given torque constraint limits after 55 interactions, whereas by using the dynamic model-based direct planning method NI-like, $\tau_{measured}$ of some points exceed the torque constraint limits, as shown in Fig. 8.

*Case 2 Cosine curve*

*Computational time*: the computational time for NI-like is 1.230 s, whereas the computational time for learning the initial trajectory is 18.322 s. However, after the learning of the initial trajectory, since the Q value has been specialized, the computational time for each learning is less than 0.1 s.

*Trajectory execution time:* the trajectory execution time by using NI-like is 1.3648 s, whereas the trajectory execution time of the initial trajectory by using TOPTO-SARSA is 1.2930s and after 8 interactions, the trajectory execution time is 1.3065s, which has a 4.2% reduction in trajectory execution time compared to use NI-like. Moreover, by using TOPTO-SARSA, $\tau_{measured}$ is no longer exceed the given torque constraint limits after 8 interactions, whereas by using NI-like, $\tau_{measured}$ of some points exceed the torque constraint limits, as shown in Fig. 9.

In conclusion, although TOPTO-SARSA requires more computational time in obtaining the initial trajectory, it can generate a more precise and safer trajectory. Moreover, by using TOPTO-SARSA to solve the TOPT problem, we do not need to build the dynamic model as well as identify the dynamic parameters (which are quite inconvenient.).

## VI. CONCLUSION

In this paper, we devote to solving the robotic time-optimal path tracking problem. Different from most of relevant researches which hope to improve the model accuracy to avoid the mismatch phenomenon, we come up with a new solution: obtaining the time-optimal trajectory only by interacting with the real world. In order to achieve this solution, we design the time-optimal path tracking problem as a reinforcement learning problem, thus the two different fields are connected. And then we propose an TOPTO-SARSA algorithm for solving the above reinforcement learning problem. By applying TOPTO-SARSA to the actual robot control through a two-step method without considering the dynamic model, it has been proved that it results in a better performance compared to the case of using dynamic model-based direct planning method. By directly interacting with the real world, the model-plant mismatch phenomenon is avoided, and the trajectory execution time obtained by using the dynamic model-free method TOPTO-SARSA is less than the trajectory execution time obtained by using model-based direct planning method. Furthermore, the actually measured torques obtained by using TOPTO-SARSA do not exceed the torque constraint limit whereas the actually measured torques obtained by using model-based direct planning method exceed the limits.

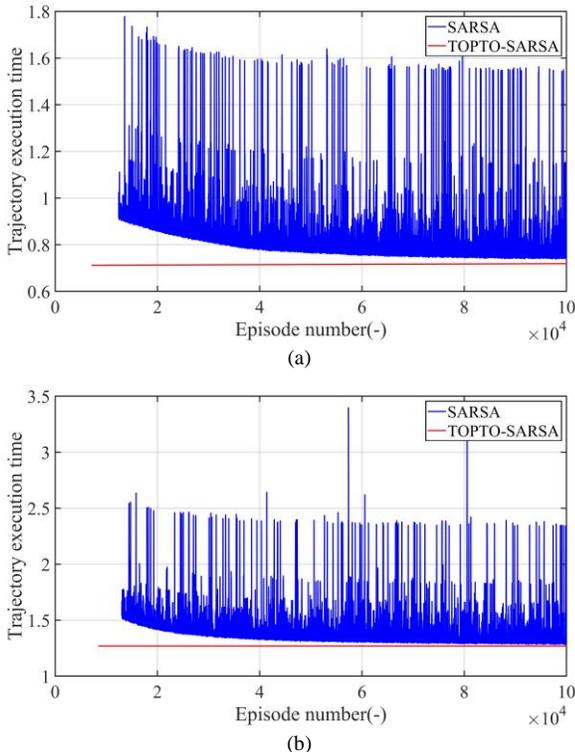

Fig. 7. Convergence analysis, where: (a) is the result of the line path and (b) is the result of the cosine curve path

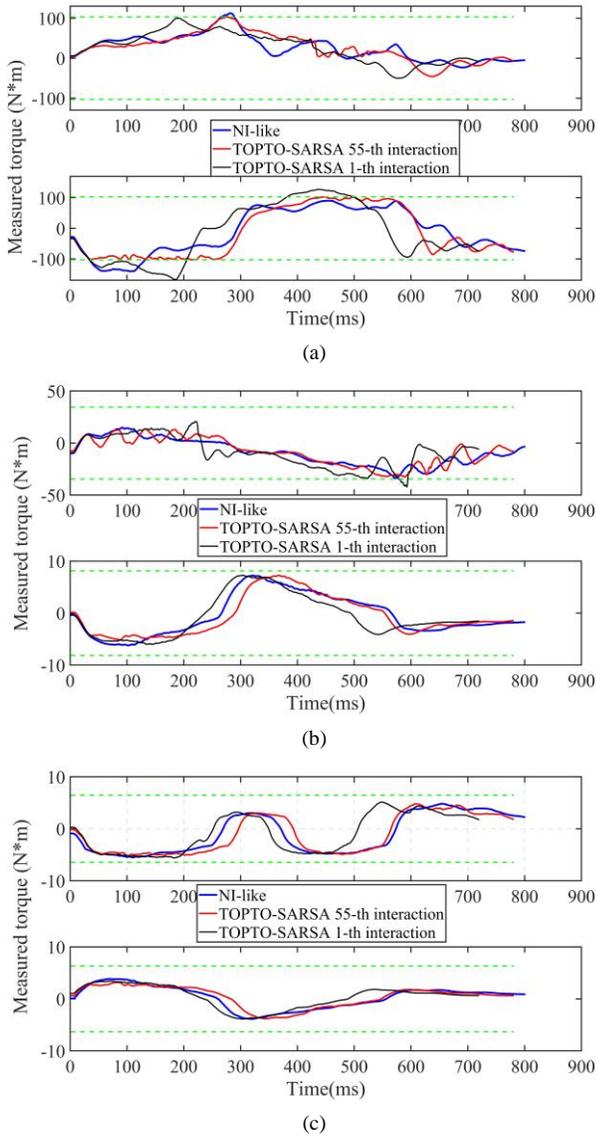

Fig. 8. Measured torques of a line path of three selected situations on all six joints of the manipulator, where (a) shows the measured torques of joint 1, 2, (b) shows torques of joint 3, 4 and (c) shows torques of joint 5, 6.

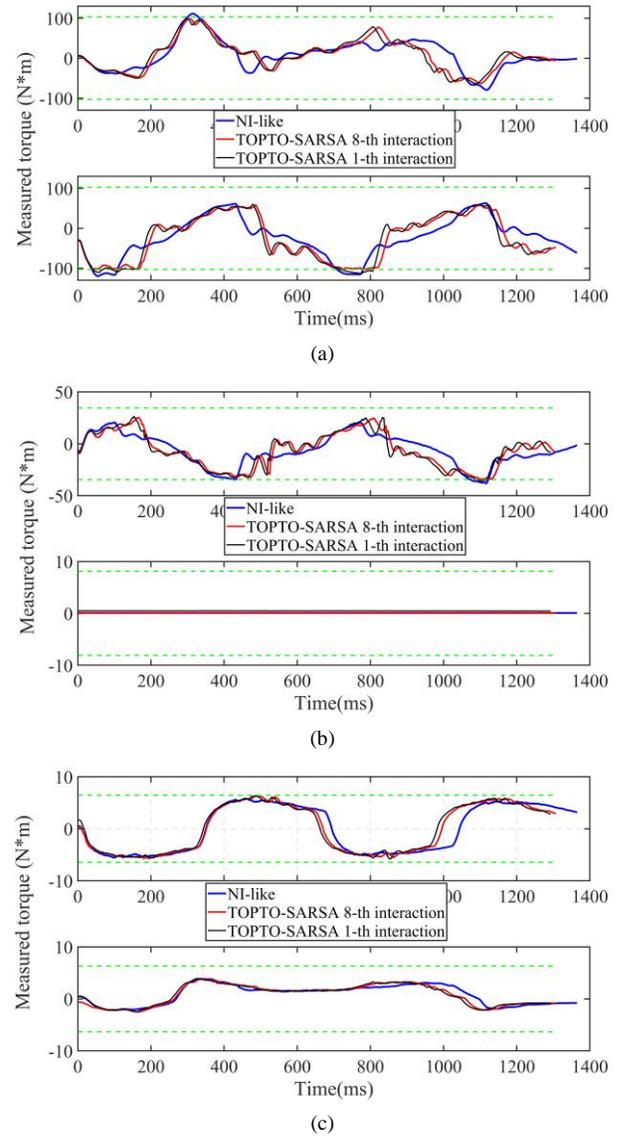

Fig. 9. Measured torques of a cosine curved path of three selected situations on all six joints of the manipulator, where (a) is the measured torques of joint 1, 2, (b) is torques of joint 3, 4 and (c) is torques of joint 5, 6.

There are several further developments which the authors intend to pursue:

1. Due to the limitation of grid, the proposed algorithms are just near optimal methods. In future, some other reinforcement learning algorithms can be considered to avoid approximation.

2. There is a significant computational time for reinforcement learning; therefore, in future, some other methods that can improve the computational efficiency can be considered.

3. Developing complex industrial scenarios for further testing of the proposed solution.

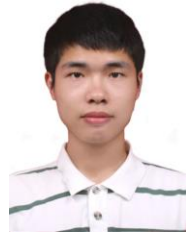

**Jiadong Xiao** received the B.S. degree in Electromechanical Engineering from Guangdong University of Techonology, Guangzhou, China, in 2017. He is currently a master candidate in School of Mechanical and Automotive Engineering, South China University of Technology, Guangzhou, China. His research interests include robot force control, robotic grinding, robotic time-optimal control.

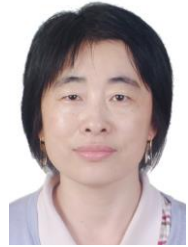

**Lin Li** received the Ph.D. degree from South China University of Technology, Guangzhou, China, in 2000. She is currently a professor of School of Mechanical and Automotive Engineering, South China University of Technology, Guangzhou. Her research interests include control technology use in industrial robot manipulators.

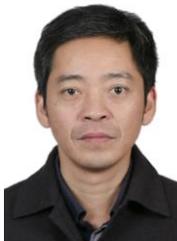

**Tie Zhang** received his Ph.D. from South China University of Technology in 2001. He is currently a professor and a Ph.D. candidate supervisor of the School of Mechanical & Automotive Engineering, South China University of Technology. He is currently the department head of Department of Mechanics and the director of Robotics Lab, South China University of Technology. His main research interests include design and control of robots, automation and intelligent systems.

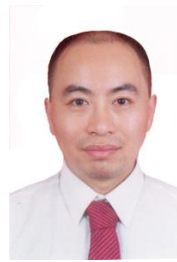

**Yanbiao Zou** received the Ph.D. degree in mechanical design and theory from South China University of Technology, Guangzhou, China, in 2005. He is an associate professor of South China University of Technology. His currently research interests include industrial robot, automation and machine learning.